\documentclass{sig-alternate}

\begin{document}


\conferenceinfo{GECCO}{2007 London, England}

\title{Robust Multi-Cellular Developmental Design}
\subtitle{{\normalsize to appear in D. Thierens et al., Eds., Proceedings of GECCO'07, ACM Press, July 2007}}



\numberofauthors{3}

\author{
\alignauthor 
Alexandre Devert\\
\affaddr{TAO / LRI ; INRIA Futurs, CNRS, Univ Paris-Sud}\\
\affaddr{F-91405, France}\\
\email{devert@lri.fr}
\alignauthor 
Nicolas Bredeche\\
\affaddr{TAO / LRI ; Univ Paris-Sud, CNRS, INRIA Futurs}\\
\affaddr{F-91405, France}\\
\email{bredeche@lri.fr}
\alignauthor 
Marc Schoenauer\\
\affaddr{TAO / LRI ; INRIA Futurs, CNRS, Univ Paris-Sud}\\
\affaddr{F-91405, France}\\
\email{marc@lri.fr}
}


\maketitle


\begin{abstract}
This paper introduces a continuous model for Multi-cellular Developmental Design. The cells are fixed on a 2D grid and exchange "chemicals" with their neighbors during the growth process. The quantity of chemicals that a cell produces, as well as the differentiation value of the cell in the phenotype, are controlled by a Neural Network (the genotype) that takes as inputs the chemicals produced by the neighboring cells at the previous time step.  In the proposed model, the number of iterations of the growth process is not pre-determined, but emerges during evolution: only organisms for which the growth process stabilizes give a phenotype (the stable state), others are declared nonviable. The optimization of the controller is done using the NEAT algorithm, that optimizes both the topology and the weights of the Neural Networks. Though each cell only receives local information from its neighbors, the experimental results of the proposed approach on the 'flags' problems (the phenotype must match a given 2D pattern) are almost as good as those of a direct regression approach using the same model with global information. Moreover, the resulting multi-cellular organisms exhibit almost perfect self-healing characteristics.
\end{abstract}



\section{Introduction}
\label{introduction}

Evolutionary Design uses Evolutionary Algorithms to design various structures (e.g.\ solid objects, mechanical structures, robots, \ldots).
It has been known for long \cite{Michalewicz} that the choice of a representation, i.e.\ of the space to search in, is crucial for the success of any Evolutionary Algorithm. But this issue is even more critical in Evolutionary Design. On the one hand, the success of a design procedure is not only measured by the optimality, for some physical criteria, of the proposed solutions, but also by the creative side of the process: a rich (i.e.\ large) search space is hence mandatory. But on the other hand, because scalability, and thus re-usability and modularity, are important characteristics of good design methodologies, the search space should have some structure allowing those properties to emerge.

The importance of the type of embryogeny (the mapping from genotype to phenotype) of the chosen representation in Evolutionary Design has been highlighted for instance in \cite{bentley-gecco-1999}, and more systematically surveyed in \cite{stanley-alife2003}. Direct representations, with no embryogeny (the relation between the phenotype and the genotype is a one-to-one mapping), have been very rapidly replaced in the history of Evolutionary Design by indirect representations, where the embryogeny is an explicit program, generally based on a grammar - and evolution acts on this program. The phenotype is then the result of the execution of the genotype. Many works have used this type of representation in Evolutionary Design, from the seminal works of Gruau \cite{gruau94genetic} and Sims \cite{sims-siggraph-1994} and their many successfull followers (cited e.g.\ in \cite{stanley-alife2003}). However, even though those representations did to some extent address the issues of modularity, re-usability and scalability, there was still room for improvement. First, the scalability is still an issue, possibly because the bigger the structure, the more difficult it is to fine-tune it through the variation operators, due to the uncontrolled {\em causality} (the effect of small mutations is not always small).
Second, the embryogeny itself, and hence the resulting structures, are not robust to perturbations \cite{bentley-naturalcomp-2005}, an important characteristic when it comes to design autonomous systems such as robots.

In order to address those issues, several recent works have chosen to use multicellular developmental models: the embryogeny is implicit, based on exchanges of some 'chemicals' between 'cells', and more or less faithfully connected to Turing's early 'reaction-diffusion' model \cite{turing-1952} (see again \cite{stanley-alife2003}, and the more recent works cited in Section \ref{relatedworks}). But several instances¹ of this model have been proposed, and a number of issues remain open, if not unsolved: 
Is the number of cells fixed, and the structure is then the result of their differentiation, or is the whole organism growing from a single cell?
Do the chemicals diffuse on a given 'substrate' or only through the interactions and exchanges among neighboring cells -- and is the topology of cell interactions fixed, evolved, or has it emerged during the development process?
What is the granularity of the possible values of chemical concentrations or quantities? When and how does development stop (the 'halting problem' of developmental approaches)? Finally, maybe the most important issue when it comes to evolve such embryogenies: what kind of 'reaction' takes place in each cell -- or, from an operational point of view, what type of controller is used within each cell, and subject to evolution? 

All those questions are of course interwined (e.g. you don't use the same type of controller depending on the type of values you intend to evolve).
However, and whatever the choices when answering the above questions, most works evolving multi-cellular developmental models report convincing results 
as far as scalability is concerned \cite{bentley-gecco-2005,federici-alife2006}, as well as unexpected robustness properties \cite{miller-eurogp-2000,bentley-naturalcomp-2005,federici-sab2006}. Indeed, even though the self-repairing capacities of the biological systems that inspired those models were one motivation for choosing the developmental approach, self-healing properties were not explicitly included on the fitnesses, and initially appeared as a side-effect rather than a target feature (see Section \ref{relatedworks} for a more detailed discussion).

This paper proposes yet another model for Multicellular Developmental Evolutionary Design. A fixed number of cells placed on a two-dimensional grid is controlled by a Neural Network. Cells only communicate with their 4 neighbors, and exchange (real-valued) quantities of chemicals. In contrast with previous works (but this will be discussed in more detail in Section \ref{relatedworks}), the phenotypic function of a cell (its type) is one of the outputs of the controller, i.e.\ is evolved together with the 'chemical reactions'. Moreover, the halting problem is implicitly left open and solved by evolution itself: development continues until the dynamical system (the set of cells) comes to a fixed point (or after a --large-- fixed number of iterations). We believe that this is the reason for the excellent self-healing properties of the organisms that have been evolved using the proposed model: they all recover almost perfectly from very strong perturbations -- a feature that is worth the additional computational cost in the early generations of evolution.

The paper is organized as follows: Section \ref{model} introduces the details of the proposed model and of its optimization using the NEAT general-purpose Neural Network evolution algorithm \cite{NEAT-2002}, that optimizes both the topology and the weights of the network. The approach is then tested in Section \ref{experiments} on the well-known 'flag' benchmark problems, where the target ``structure'' is a 2D image. A meaningful validation is obtained by comparing the results of the developmental approach to those of the data-fitting approach: the same neural optimization method is used but the inputs are the coordinates (x,y) of the cell: indeed, it should not be expected to obtain better results with the developmental approach than with this direct data-fitting approach. Furthermore, the excellent self-healing properties of the resulting structures are demonstrated. Those results are discussed in Section \ref{relatedworks} and the proposed approach is compared to other existing approaches for Multicellular Developmental Design. Finally, further directions of research are sketched in concluding Section \ref{conclusion}.

\section{Developmental model}
\label{model}
The context of the proposed approach is what is now called  Multi-Cellular Development Artificial Embryogeny \cite{federici-ppsn2004}: An \textbf{organism} is composed of identical \textbf{cells}; Each cell encapsulates a \textbf{controller} (loosely inspired from a biological cell's regulatory network); 
All cells, and thus the organism, are placed in a \textbf{substrata} with a given topology; Cells may eventually divide (i.e. create new cells), differentiate (i.e. assume a predefined function in the phenotype), migrate and/or communicate with one another in the range of their neighborhood.

In the literature there is a clear distinction between approaches that do not rely on cell division, and thus require that the environment is filled with cells at startup \cite{bentley-naturalcomp-2005}, and approaches where cells divide and migrate \cite{federici-worlds2004,miller-gecco-2004}. In both case however, communication may be performed from one cell to another \cite{bentley-naturalcomp-2005,federici-worlds2004}  (direct \textbf{cell-cell} mechanism) or diffused through the environment \cite{miller-gecco-2004} (substrata diffusion mechanism of \textbf{chemicals}).

A cell or group of cells ``grows'', or ``develops'', by interacting with the environment, usually at discrete time steps. This process stops at some point and the organism is evaluated w.r.t.\ the target objective. In all the works that are cited above, the growth stop is forced (development is stopped after a predefined number of steps). Defining an efficient endogenous stopping criterion can be related to addressing the \textbf{halting problem} for Developmental Embryogeny. 


In this context, the model proposed in this paper has the following characteristics: a fixed number of cells are positioned on a two-dimensional non-toroidal array (no cell division or migration). The state of each cell is a vector or real values, and the controller is a Neural Network. Cells produce a predefined number of 'chemicals' that diffuse by a pure cell-cell communication mechanism. Time is discretized, and at each time step, the controller of each cell receives as inputs the quantities of chemicals produced by its neighboring cells (4-neighbors Von Neuman neighborhood is used - boundary cells receive nothing from outside the grid). The neural controller takes as external input the chemicals of the neighboring cells and  computes a new state for the cell, as well as the concentrations of the chemicals to be sent to neighboring cells at next time step. No global information is available or transmitted from one cell to another -- the challenge is to reach a global target behavior from those local interactions.

As noted in the introduction, this model can be thought of as a simplified instance of Turing's reaction-diffusion model \cite{turing-1952}, with discretized time and space. But it can also be considered as a very simple model of a Genetic Regulatory Network \cite{banzhaf-ecal2003}. The topology of the network is fixed, all 'genes' produce the same 'proteins', but the activation/inhibition of protein production is given by the (non-linear) neural network function.
Finally, looking beyond biological analogies, the proposed model can also be seen as a Continuous Cellular Automata \cite{Wolfram}, i.e.\ cellular automata with continuous states and discrete time, more precisely as a Cellular Neural Network \cite{CNN88}, cellular automata where the update rule for each cell is given by a neural network, typically used in VLSI design. 

\subsection{The Neural Network Controller}

In this work, the state of a cell, that is responsible for for both its differentiation (i.e.\ its phenotypic expression) and the communication with other cells though the diffusion of chemicals, is a vector of real values:  a single real value (gray level) in the 'flag' applications described in Section \ref{experiments} -- though more complex environments could require more complex differentiation states. Hence the widely used and studied model of Discrete Time, continuous state, Recurrent Neural Network (DTRNN) with sigmoidal transfer functions was chosen for the cell controllers.
This choice of a Neural Network as a controller of the cells was inspired by the long-known property that Neural Networks are Universal Approximators \cite{Hornik89}. 
The inputs of the Neural Network are the values of the chemical quantities coming from the 4 neighbors of the cell. Its outputs are the state of the cell plus one output per chemical. If there are $N$ neurons and $M$ external inputs, the more general form of update rule at time step $t$ for neuron $i$ of a DTRNN is 

\[
a_i(t+1) =
\sigma(
  \sum_{j=1}^{N} w_{i,j} a_j(t) +
  \sum_{j=1}^{M} z_{i,j} I_j(t)
)
\]
{\bf En général, l'activation est juste la somme pondérée des entrées. Ce que tu donnes est la sortie. De plus, IL FAUT DISTINGUER LES ESPECES ???}
where $a_i(t)$ is the activation of neuron $i$ at time $t$,
 $I_j(t)$ is the $j^{th}$ external input at time $t$, 
$w_{i,j}$ is the weight of the connection from neuron $j$ to neuron $i$ (0 if no connection exists), 
$z_{i,j}$ is the weight of the connection from input $j$ to neuron $i$, and 
$\sigma(x) = \frac{1}{1 + e^{-x}}$ is the standard sigmoid function.

It is important to note that, even if the neural controller is a feedforward neural network (i.e.\ there are no loops in the connection graph), the complete system is nevertheless a large recurrent neural network because the exchanges of chemicals between the cells do create loops. In this respect, the chemicals can be viewed as an internal memory of the whole system. Figure 1 
shows a schematic view of a cell with its 4 neighbors, that uses two chemical concentrations to communicate. The cell transmit the same concentrations of each chemical to it neighboring cells, so we have only 2 outputs but 8 inputs. An additional output (not shown) is used for the differentiation value.


Obviously, this model can be easily extended to any number of
chemicals, as well as to any dimensions for the state of the cells,
allowing differentiations into more sophisticated mechanical parts
(e.g.\ robot parts, joints with embedded controller, etc).

\begin{figure}
\begin{center}
\epsfig{file=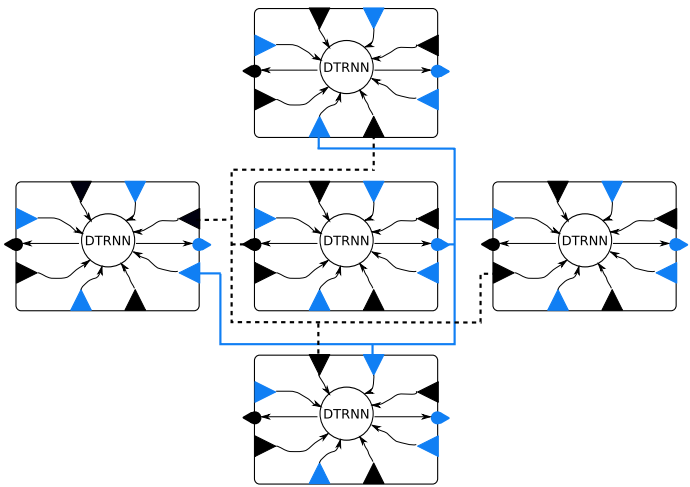, height=2.5in}
\caption{Schematic view of teh connections between cells in the case of 2 chemicals}
\end{center}
\label{fig:controler}
\end{figure}





\subsection{Controller Optimization}

Even though the smaller class of simple sigmoidal 3-Layer Perceptron has the Universal Approximator property, determining the number of hidden units for a MLP remains an open issue, and practical studies have demonstrated that exploring the space of more complex topologies (including recurrent topologies) could be more efficient than just experiencing with a one hidden layer perceptron. 
Moreover, many algorithms have been proposed for the evolution of Neural Networks, and a good choice for the evolution of cell controllers was the \emph{NEAT} algorithm \cite{NEAT-2002}, a state-of-the-art evolutionary NN optimization algorithm that makes it possible to explore both feedforward and recurrent topologies.

This algorithm relies on a direct encoding of neural network topologies that are evolved using a classical evolutionary stochastic optimization scheme. The main feature of NEAT is that it explores the topologies from the bottom-up: starting from the simplest possible topology for the problem at hand, it performs variations over individuals by adding neurons and connections to networks in such a way that the behavior of the network is preserved at first - this makes it possible to explore topology in a non destructive fashion.


Our NEAT implementation has been validated from published results. For all the experiments in this paper, NEAT parameters have been set to the values given in \cite{NEAT-2002} for solving the sample \emph{XOR regression} and \emph{double-pole balancing} tasks. Those values seemed robust for the problem at hand, according to a limited parametric study. They are summarized in table \ref{tab:NEAT-parameters}. 

\begin{table}
\centering
\begin{tabular}{|c|c|}
\hline
Population size & 500 \\
\hline
Max. number of evaluations & 250000 \\
\hline
Reproduction ratio per species & 0.2 \\
\hline
Elite size per species & 1 \\
\hline
Crossover prob. & 0.15 \\
\hline
Add-node mutation prob. & 0.01 \\
\hline
Add-link mutation prob. & 0.01 \\
\hline
Enable-link mutation prob. & 0.045 \\
\hline
Disable-link mutation prob. & 0.045 \\
\hline
Gaussian weights mutation prob. & 0.8 \\
\hline
Std. dev. for Gaussian weight mutation & 0.1 \\
\hline
Uniform weights mutation prob. & 0.01 \\
\hline
Distance parameters for fitness sharing & 1.0 --  1.0 -- 0.2 \\
\hline
\end{tabular}
\caption{NEAT parameters (see [17] for details).} 
\label{tab:NEAT-parameters}
\end{table}

As already noted, an interesting feature of NEAT algorithm is that it can handle the evolution of both feedforward and recurrent neural networks -- hence allowing an easy comparison of both models. Another interesting feature of NEAT is that it allows the user to declare some constraints on the topology - in this case, all input and output neurons are forced to be connected to at least one neuron in the controller.

\subsection{Halting the Growth Process}
\label{halting}
In Multi-cellular developmental systems, the phenotype (the target structure to be designed, on which the fitness can be computed) is built from the genotype (the cell-controller, here a Neural Network) through an iterative process: 
Starting from a uniform initial condition (here, the activity of all neurons is set to 0), all cells are synchronously updated, or, more precisely, all neurons of all cells are synchronously updated, in case the neural network is recurrent. But one major issues of such iterative process is to determine when to stop.

In most previous approaches (see Section \ref{relatedworks}), the number of iterations is fixed once and for all by the programmer. However, this amounts to adding one additional constraint to the optimization process: Indeed, it is clear that the number of iterations that are necessary to reach a given state depends on that state, but also on the organism. Moreover, it also most probably should depend on the conditions of the experiment: the dimension of the grid, the number of chemicals, \ldots 

Because there seems to be no general way to a priori determine the number of iterations that should be allocated to the organisms to reach a target phenotype, a good solution is probably to leave this parameter free, and to let evolution tune it.

One straightforward way to do so would be to compute the fitness of the organism at all stages of the iterative process, i.e.\ on all intermediate states of the cells. However, because such computation might be very heavy (for instance when designing mechanical structures, one often has to compute their fitness using some FEM analysis) this solution has been rejected. On the other hand, if we suppose that cell updates are cheap to compute compared to the actual fitness of a phenotype, it is possible to let the system iterate until it stabilizes. Of course, as is known from the Cellular Automata point of view \cite{Wolfram}, some systems will never stabilize, having either a chaotic behavior, or approaching some non-stationary attractor. However, one can hope that the set of systems that actually do reach a fixed point is rich enough to contain good solutions to the problem at hand.

The next challenge is to detect when the system stabilizes. It is
proposes here to compute some {\em energy} of the system at each
iteration, and to stop when this energy remains constant
during a certain number of iterations.  

More precisely, the energy of the system is computed as the sum of the activations of all neurons of all cells:
\[
E(t) = \sum_{all neurons} a(t)^2,
\]

and the organism is considered stable when $E(t)= E(t+1)$ 
during a given number of time steps. Of course, a maximum number of
iterations is given, and a genotype that hasn't converged after that
time receives a very bad fitness: such genotype has no
phenotype, so the fitness cannot even be computed anyway. 
After such a final stable state for the organism has been reached, it
is considered as the phenotype and undergo evaluation.


\section{Experiments}
\label{experiments}
Even though the long term goal of Developmental Design is to design mechanical structures (bridges, buildings, robots, \ldots), the computational cost of mechanical simulations makes such applications out of reach at the moment. Moreover, the classical benchmarks that have been used to evaluate developmental approaches in recent works is the \emph{flag problem}, as originally proposed by Miller \cite{miller-french-flag} and later used by other researchers in the field \cite{federici-worlds2004,bentley-gecco-2005}.

The cells are the square tiles of a
rectangular grid, like the pixels of a digitalized picture. At the end
of the developmental phase, the cells must differentiate into a (generally discrete) color state so that the whole organism matches a
given target picture. The rule of the game is that cells should of course have no access to information about the target picture, or to global informations like their absolute position in the grid. 
The only feedback from the target is the fitness of the 
phenotype, provided by a similarity measure. Pictures with simple
patterns remains the most widely used, like the French or Norwegian flags.

\subsection{On Fitness and Flags}
\label{fitnessFlag}
The problem is to define a similarity measure between the final state of the cells on the grid after differentiation (i.e. each cell has a color) and the target flag. While most previous works use discrete states
as color values, with 3 or 4 different states, the continuous values taken by the cell states in the Neural Network model allow a more precise sampling of the possible colors. Hence all experiments reported in the following use grayscale pictures with 256 gray levels: the output of the controller, that is in [0,1], is discretized, and the fitness is computed on the discrete values as follows. 

The fitness measure is the similarity between the picture generated by 
the developmental process and the target picture. A smooth similarity $s(A,B)$ between two pictures $A$ and $B$ with both $w \times h$ pixels is defined by:  
\[
s(A,B)= \frac{1}{wh} \sum_{i=0}^{h-1} \sum_{j=0}^{w-1} (A(i,j)-B(i,j))^2
\]

The value $s(A,B)$ lies in $[0,1]$, and reaches 1 if $A=B$.

\begin{figure}
\centering
\begin{tabular}{cccc}
  \epsfig{file=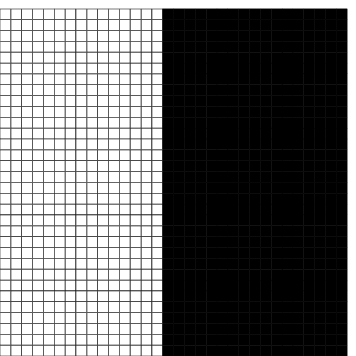, height=0.7in} & 
  \epsfig{file=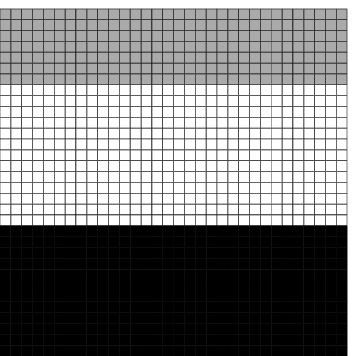, height=0.7in} & 
  \epsfig{file=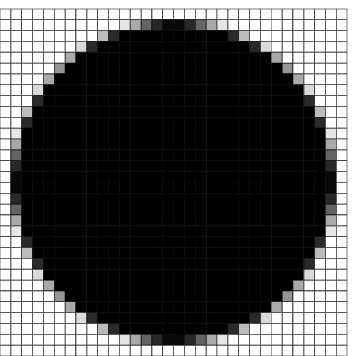, height=0.7in} & 
  \epsfig{file=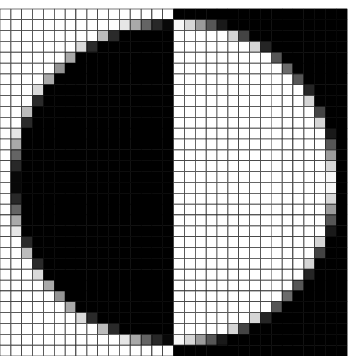, height=0.7in} \\
2-bands & 3-bands & disc & half-discs
\end{tabular}
\caption{The four target pictures}
\label{theFlags}
\end{figure}

Four $32 \times 32$ target pictures are used, that can be seen on Figure \ref{theFlags}. The first picture is a simple 2-bands symmetrical picture with 2 colors. The 3-bands image contains 3 horizontal layers of different colors, and should be slightly more difficult to retrieve. However, because of the x-y bias of the chosen representation (information is transmitted horizontally or vertically), the last 2 images should be (and will be) more difficult to grasp, as they contain circular patterns. Note that because scalability is not the primary issue under study in this work, only pictures of this medium size (compared to previous work on the flag problem) are used.

\subsection{The Different Models}
In order to explore different models within the general context described in Section \ref{model}, 4 instances of the proposed model are experimented with: 
a feedforward neural network, and 1 chemical, termed \emph{1-ffwd}, a recurrent neural network with 1 chemical, termed \emph{1-recurr}, a recurrent neural network with 2 chemicals, termed \emph{2-recurr}, a feedforward neural network, and 2 chemicals, termed\emph{2-ffwd}.

However, there are (at least) two possible causes of error in the proposed approach: on the one hand, there might not exist any fixed point of the multi-cellular developmental systems under study that can approximate the target image; but on the other hand, even if a good solution does exist, the chosen computation method (evolutionary optimization of a neural network using NEAT) might not be able to approximate it. Note that this situation is common to all computational approaches of complex systems: the former error is termed 'modeling error' and the latter 'method error'. A third type of error is also reported in numerical experiments, the 'numerical error', due to propagating round-offs, and will be neglected here.

In order to try to discriminate between the modeling error and the method error, a fifth model is also run, on the same test cases and with similar experimental conditions than the four developmental approaches described above: the layout is exactly the same (a 2D grid of cells), the same NEAT parameters are used (to evolve a feedforward neural network), and selection proceeds using the same fitness. However, there is no chemical nor any exchange of information between neighboring cells, and on the other hand,  all cells receive as inputs their (x,y) coordinates on the grid. Hence the flag approximation problem is reduced to a simple regression problem. In the following, the results of this model will be considered as reference results, as it is not expected that any developmental approach can ever beat a totally informed model using the same NEAT optimization tool. This experiment is termed ``\emph{f(x, y) = z}`` from now on.

\subsection{Experimental setup}

All 5 models described in previous section have been run on the 4 flags showed on Figure \ref{theFlags}. All results presented in the following are statistics over 16 independent runs. 

As already said, the evolutionary 
neural network optimizer is NEAT, with the settings that are described in
Table \ref{tab:NEAT-parameters}. It is worth noticing that during all runs, no bloat was ever observed for the NEAT genotypes. The mean size of the networks (measured by the total number of edges between neurons) gently grew from its starting value (between 5 and 10 depending on the model) to some final value below 40 -- the largest experiment reaching 45. This first result confirms the robustness of this optimization tool, but also, to some extent, demonstrates the well-posedness of the problems NEAT was solving (bloating for Neural Networks can be a sign of overfitting ill-conditioned data).

As argued in section \ref{halting}, the halting of the growth process is based on  the stabilization of the energy of the organism, checked over some time window. 
The width of this time window has been set to 8 time-steps in all experiments. However, because not all networks will stabilize, a maximum number of iterations has to be imposed. This maximum number was set to 1024, and if no stabilization has occurred at that time, the fitness is set to value 0: as 0 is the worst possible value for the fitness, this amounts to using some death penalty for the stabilization constraint.

With such settings, a typical run lasts about one day on a 3.4GHz Pentium IV. Though this might seem a huge computational cost, we believe that it is not a critical issue when designing real-world structures: On the one hand, designing mechanical parts is already a time-consuming process, involving highly trained engineers -- and human time nowadays costs much more than CPU time. On the other hand, when the structure that is being designed is bound to be built by thousands or millions, a few days represent a very small overhead indeed.



\begin{figure}
\centering
\epsfig{file=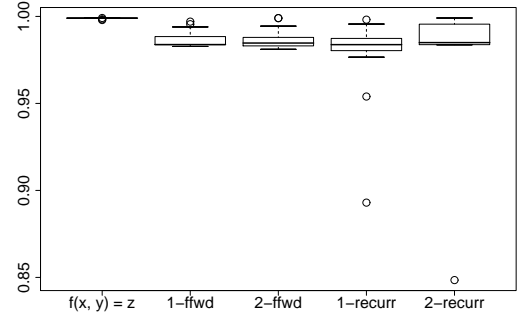, height=2.8in, angle=270}
\caption{Off-line results for the 3-bands problem}
\label{fig:three-bands-stats}
\end{figure}

\begin{figure}
\centering
\epsfig{file=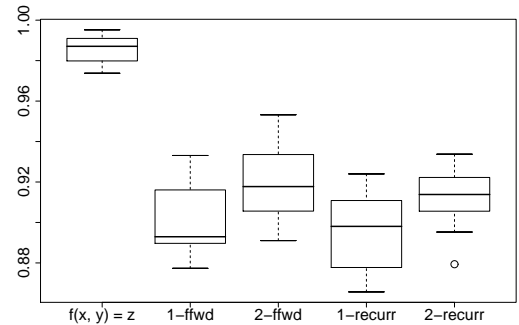, height=2.8in, angle=270}
\caption{Off-line results for the disc problem}
\label{fig:disc-stats}
\end{figure}

\subsection{Results}

\subsubsection{Comparing fitnesses}

\begin{figure}
\centering
\epsfig{file=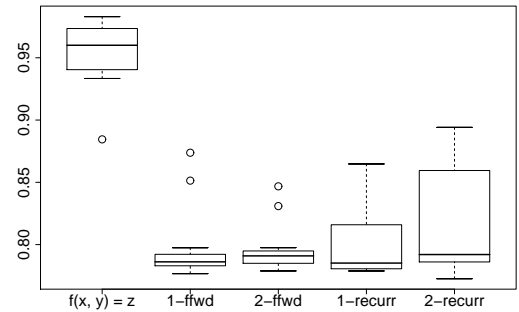, height=2.8in, angle=270}
\caption{Off-line results for the half-discs problem}
\label{fig:half-discs-stats}
\end{figure}

\begin{figure}
\centering
\epsfig{angle=270,width=7cm,file=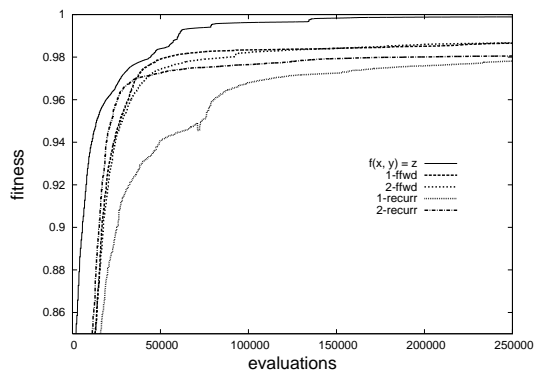}
  \caption{Evolution of average of best fitness for the 3-bands problem. The lowest curve is that of the {\it 1-recurr model}, and the 3 indistinguishable curves above the other 2 are those of the other 3 artificial embryogeny models.}
  \label{fig:evolution-three-bands}
\end{figure}

The statistics for the off-line results are displayed as the usual box-plots\footnote{as generated by the R statistical package, see  {\tt http://en.wikipedia.org/wiki/Box\_plot} for a precise description.} on Figures \ref{fig:three-bands-stats}, \ref{fig:disc-stats} and \ref{fig:half-discs-stats} respectively for the 3-bands, disc and half-discs problems of Figure \ref{theFlags}, and on-line results (the average over the 16 runs of the fitness of the best-of-generation individuals as evolution progresses) are shown on Figure \ref{fig:evolution-three-bands} for the 3-bands problem.



The results for the 2-bands problem are almost identical for the 5 models, and are not presented here: same average fitness of 0.999, with a slightly larger variance for the developmental approaches (and variance 0 for the regression model). For each setting of the embryogenic approach, though, some runs were able to find a marginally better solution than that of the regression model -- but without any statistical significance. For the slightly more difficult target \emph{three-bands}, the reference model is still able to find an exact solution, as shown in figure \ref{fig:three-bands-stats}, while the 3 embryogenic models give nearly optimal individuals.


As expected, the \emph{disc} target is difficult for the embryogenic approaches: as can be seen on the box-plots (Figure \ref{fig:disc-stats}, 
all 4 are clearly outperformed by the reference model, that was not trapped in the same local optimum. The on-line results did not reveal any other conclusion, and are not shown here. It is worth noting that here, experiments using 2 chemicals outperform the same  model with a single chemical (with statistically significant differences according to a 95\% confidence T-test).

Finally, the situation is slightly different for the half-discs, the most difficult target (Figure \ref{fig:half-discs-stats}): all embryogenic models are, again, clearly outperformed by the reference model, even though this model doesn't reach such a good fitness than for the disc problem. However, the best results among embryogenic approaches are obtained by the recurrent networks, that exhibit a much larger variance, and thus sometimes reaches much better fitnesses -- with a slight advantage for the 2-chemicals recurrent model in this respect.

\subsubsection{Halting Criterion and Robustness}
\label{healing}

\begin{figure}
\centering
  \begin{tabular}{c|ccc}
    neuron 0 (gray level)&
    \epsfig{file=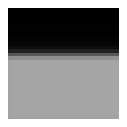}& 
    \epsfig{file=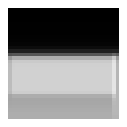}& 
    \epsfig{file=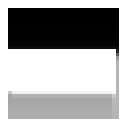}\\
    \hline
    neuron 1 (chemical 1)&
    \epsfig{file=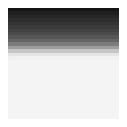}&  
    \epsfig{file=pictures/three-bands-sequence/chemical-1-epoch-00016.eps}&
    \epsfig{file=pictures/three-bands-sequence/chemical-1-epoch-00016.eps}\\
    \hline
    neuron 2 (chemical 2)&
    \epsfig{file=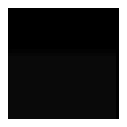}&
    \epsfig{file=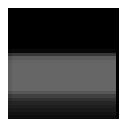}&
    \epsfig{file=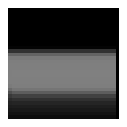}\\
  \end{tabular}
  \caption{Development stages on the three-bands problem for the recurrent NN with 2 chemicals at iterations 16, 32 and 44 (columns) for the phenotype (top row), and both chemicals.}
  \label{fig:development-example}
\end{figure}

\begin{figure}
\centering
  \begin{tabular}{c|ccc}
    neuron 0 (gray level)&
    \epsfig{file=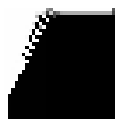}& 
    \epsfig{file=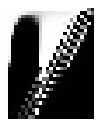}& 
    \epsfig{file=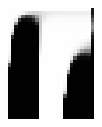}\\
    \hline
    neuron 1 (chemical 1)&
    \epsfig{file=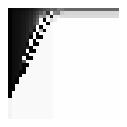}&  
    \epsfig{file=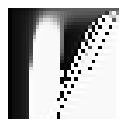}&
    \epsfig{file=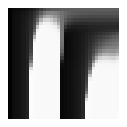}\\
    \hline
    neuron 2 (chemical 2)&
    \epsfig{file=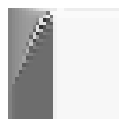}&
    \epsfig{file=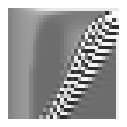}&
    \epsfig{file=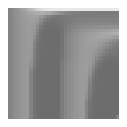}\\
  \end{tabular}
  \caption{Development stages on the half-discs problem for the recurrent NN with 2 chemicals at iterations 28, 64 and 122 (columns) for the phenotype (top row), and both chemicals.}
  \label{fig:development-half-discs}
\end{figure}

The evolved halting criterion is one of the main original feature of the proposed approach. It thus needs to be studied in detail, especially as it is closely related to the self-healing properties, i.e. the robustness with respect to noise during the growth iterations.

Because all organisms are allowed 1024 iterations in their growth process, it could be feared that several hundreds iterations would be needed before stabilization even for the best solutions found by the algorithm. The total computational costs would henceforth have been tremendously higher that it already is. The good news is that in all cases, and for all embryogenic models, the whole population rapidly contains a large majority of organisms that did stabilize within a few dozens iterations. Illustrations of the growth process are given in Figures \ref{fig:development-example} and  \ref{fig:development-half-discs}. For the easy 3-bands problem, only 44 iterations are needed (and chemical 1 doesn't change after the 16th iteration). For the more difficult half-discs problem, 122 iterations are needed. 

But another important issue is that of robustness: earlier works \cite{miller-gecco-2004,federici-sab2006} have  demonstrated that developmental approaches lead to robust solutions as far as development is concerned 

\begin{figure}
\centering
  \begin{tabular}{ccccc}
    \epsfig{file=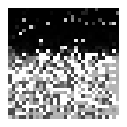} & 
    \epsfig{file=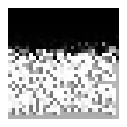} & 
    \epsfig{file=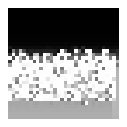} &
    \epsfig{file=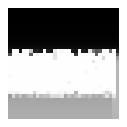} &
    \epsfig{file=pictures/three-bands-sequence/epoch-00044.eps}
  \end{tabular}
  \label{tab:self-healing-example}
  \caption{Self-healing on the three-bands problem for the recurrent NN and 2 chemicals: Snapshots of the phenotype at iterations 0 (beginning of the perturbation), 4, 11, 17 and 22.}
\end{figure}

Here, the robustness of the fixed points was checked by applying a centered Gaussian
perturbation with unit standard deviation to the states of all neurons. The good news is that for any perturbation, 
100\% of the feedfoward controllers and 75\% of the recurrent
controllers return to the very same state they had before the
perturbation. The other 25\% reccurent controllers returns to a state
very close to the one they had before perturbation. An example of perfect
and fast self-healing for the \emph{three-bands} problem is shown in
Figure \ref{tab:self-healing-example}.


To sum up, the embryogenic approach perform often nearly as good as
a simple regression (\emph{f(x, y) = z}), if using the same
optimizer. The feedforward and the recurrent networks seem hardly distinguishable across the 4 experiments, and a slight advantage of the 2-chemical over the 1-chemical could be hypothesized. The most interesting result concerns the almost perfect self-healing property of the resulting organisms.



 ~\\~\\ 
 
\section{Related Works and Discussion}
\label{relatedworks}

This section discusses the proposed approach in the light of other works on  multi-cellular embryogenies from the litterature. 

The pioneering work by Julian Miller \cite{miller-gecco-2004} belongs to the 'duplicating cells' category: Cells are allowed to duplicate, and growth starts with a single cell. Cells achieve communication by placing chemicals at their location and reading chemicals from their 8 neighbors. Moreover, a hand-written mechanism ensures their diffusion on the grid. Each cell can also differentiate into one of four cell types (one of the three colors, or the 'dead cell' tag) and each cell communicates its type to neighboring cells. The cell controller is designed as a boolean logic circuit optimized with Cartesian Genetic Programming \cite{miller-eurogp-2000} and the task is to find an organism that fits a 12x9 French flag. Experiments are conducted with a varying number of chemicals (from 0 to 4) and results showed that after 10  iterations, the french flag could be reproduced with nearly 95\% similarity. Even more interesting results concerning self-repairing showed that with varying perturbations, the system could still recover and converge toward patterns that are somewhat similar (though not identical) to the ones it would have achieved without perturbations.

In \cite{federici-worlds2004}, Diego Federici extends Miller's work: again, only one single cell exists at iteration 0, and duplication is allowed. Each cell gets as input the 4 neighboring cell types and one single chemical concentration, resulting here also from a hand-written diffusion rule. The controller is a multi-layer perceptron with fixed topology (only the weights are optimized) and the task is to fit a set of 9x6 flags (including the Norwegian flag). One interesting feature is that the optimization process is twisted to favor diversity, and implements a clever problem decomposition scheme named "multiple embryogenic stages"  with convincing results.

The work by Gordon and Bentley \cite{bentley-naturalcomp-2005} differs from previous approaches by considering only communication and differentiation in the substrata. The grid starts with a cell at all available grid points, and cells communicate by diffusing chemicals to neighboring cells only. Each cell then receives as input one chemical concentration, computed as the average of the concentrations of all neighboring cells: hence, no orientation information is available. In the Cellular Automata context, such system is called a totalistic automaton. One drawback of this approach is that it requires that some cells have different chemicals concentration at start-up. Furthermore, it makes the whole model biased toward symmetrical patterns ("four-fold dihedral symmetry"). The controller is a set of 20 rules that produce one of the four chemicals and sends it towards neighboring cells. The set of rules is represented by a bit vector and is evolved using a classical bitstring GA. The paper ends with some comparisons with previous works, namely  \cite{federici-worlds2004,miller-gecco-2004}, demonstrating comparable and sometimes better results. But a possible explanation for that success could be the above-mentionned bias of the method toward symmetrical patterns.

The approach proposed here shares some similarities with the approaches described above. The controller is defined as a neural networks, as in \cite{federici-worlds2004}; but in contrast to \cite{federici-worlds2004}, both the topology and the weights are optimized, thanks to NEAT. Further work should determine whether this difference is essential or not by running the same algorithm (i.e. with the stabilization incentive in the fitness) and multi-layer perceptron controllers.

However, there are even greater similarities between the present work and that in \cite{bentley-naturalcomp-2005}. In both works, the grid is filled with cells at iteration 0 of the growth process (i.e. no replication is allowed) and chemicals are propagated only in a cell-cell fashion without the diffusion mechanisms used in \cite{federici-worlds2004,miller-gecco-2004}. Indeed, a pure cell-cell communication is theoretically sufficient for modelling any kind of temporal diffusion function, since diffusion in the substrata is the result of successive transformation with non-linear functions (such as the ones implemented by sigmoidal neural networks with hidden neurons). However, this means that the optimization algorithm must tune both the diffusion reaction and the differentiation of the cells.
On the other hand, whereas  \cite{bentley-naturalcomp-2005} only consider the average of the chemical concentrations of the neighboring cells (i.e. is {\em totalistic} in the Cellular Automata terminology), our approach does take into account the topology of the organism at the controller level, de facto benefitting from orientation information. This results in a more general approach, though probably less efficient to reach symmetrical targets. Here again, further experiments must be run to give a solid answer.

But two main issues contribute to the originality of the approach proposed here: (1) the output for cell differentiation is a continuous value, and (2) the halting problem is indirectly addressed through the fitness function, that favors convergence towards a stable state (i.e.\ a fixed point). 

Indeed, all other works consider that a cell may differentiate into one of a given set of discrete states (e.g. blue, red, and white) while output is considered here as a continuous value (discretized into a 256-gray level value). At first sight, this can be thought as making the problem harder by increasing the size of the search space. However, it turns out that a continuous output results in a rather smooth fitness landscape, something that is known to be critical for Evolutionary Algorithms. Additional experiments (not reported here) did demonstrate that it was much harder to solve the same flag problems when discretizing the controller outputs before computing the fitness (Section \ref{fitnessFlag}). Indeed, discretized outputs lead to a piecewise constant fitness landscape and the algorithm has no clue about where to go on such flat plateaus. However, here again, more experiments are needed before drawing strong conclusions.

Secondly, from a dynamical system viewpoint, the objective function can be seen as selecting only the organisms that do reach a fixed point, starting from given initial conditions defined as uniformly initialized cells. First, all previous works needed to a priori decide the number of iterations that growth would use, and it is clear that such parameter is highly problem dependent, and hence should made adaptive if possible.
But more than that, the good news is the strength of the fixed point reached by the organism, its attracting power when starting from other initial conditions - that is, an extreme case of self-healing capabilities against perturbations for the organism. Of course, previous works \cite{miller-gecco-2004} already noted the growth process is remarkably stable under perturbations, and is able to reach a pattern quite similar (though not identical) to the original target pattern. However, it should be noted that the organisms evolved in \cite{miller-gecco-2004} keep on growing if growth is continued after the fixed number of iterations, and eventually turn out to completely diverge from the target pattern. Similarly, \cite{federici-worlds2004} observes that a perturbation in the earlier stages of development leads to an increase in the disruption of the final pattern that is linear with respect to the number of development steps. Robustness towards perturbation was later confirmed and more thoroughly studied in \cite{federici-sab2006}.

But, as demonstrated by the experiments shown in Section \ref{healing}, our model achieves astounding results regarding the self-healing property. Starting from completely random conditions (i.e.\ inputs and outputs set to random values), the system is able to perform a 100\% recovery and to converge to the exact pattern that was reached during evolution (i.e. when starting with value 0 for all neuron activations). Some runs were performed without the stabilization criterion, the final individuals never shown such properties. This seems to be a clear consequence of the way stabilization is favored in the fitness function -- though the precise reason for the extraordinary absorbing property of all fixed points reached in the experiments so far remains to be understood. 







\section{Conclusion}
\label{conclusion}

This paper has introduced a continuous Neural Network model for Multi-Cellular Developmental Design. The Neural Network is evolved using the state-of-the-art NEAT algorithm that optimizes both the topology and the weights of the network, and can evolve both feedforward and recurrent neural networks. The model was validated on four instances of the 'flag' problem, and on 3 out of 4 instances it performed as good as NEAT applied to the equivalent regression problem: this is a hint that the modeling error of the developmental approach is not much bigger than that of the Neural Network approach for regression (which is proved to be small, thanks to the Universal Approximator property), and is in any case small compared to the computational error (i.e. the error done by NEAT when searching the globally optimal network).

But the most salient feature of this model lies in the stopping criterion for the growth process: whereas most previous work required to a priori decide on a number of iterations, the proposed algorithm selects organisms that reach a fixed point, making the stopping criterion implicitly adaptive. The major (and somewhat unexpected) consequence of this adaptivity is the tremendous robustness toward perturbations during the growth process: in almost all experiments, the fixed point that is reached from the initial state used during evolution (all neural activations set to 0) seems to be a global attractor, in the sense that the organism will end up there from any starting point.

\bibliographystyle{abbrv}
\bibliography{biblio}


\end{document}